\documentclass{ecai2000}
\usepackage{times}
\usepackage{graphicx}
\usepackage{latexsym}
\usepackage{macros}   

\begin{document}

\title{Naive Bayes and Exemplar-Based approaches to Word Sense
Disambiguation Revisited}

\author{Gerard Escudero, Llu\'{\i}s M\`arquez\ {\rm and}\
German Rigau\institute{TALP Research Center, Software 
Department, Technical University of Catalonia, Jordi Girona Salgado 1-3, 
Barcelona E-08034, Catalonia, email: \{escudero, lluism, g.rigau\}@lsi.upc.es}}

\maketitle
\bibliographystyle{ecai2000}

\begin{abstract}
This paper describes an experimental comparison between two standard 
supervised learning methods, namely Naive Bayes and Exemplar--based 
classification, on the Word Sense Disambiguation (\aWSD) problem. 
The aim of the work is twofold.
Firstly, it attempts to contribute to clarify some confusing information 
about the comparison between both methods appearing in the related 
literature. In doing so, several directions have been explored, 
including: testing several modifications of the basic learning 
algorithms and varying the feature space. 
Secondly, an improvement of both algorithms is proposed, in order 
to deal with large attribute sets.
This modification, which basically consists in using only the 
{\it positive} information appearing in the examples, 
allows to improve greatly the efficiency of the methods, 
with no loss in accuracy. 
The experiments have been performed on the largest sense--tagged 
corpus available containing the most frequent and ambiguous English 
words. Results show that the Exemplar--based approach to \aWSD\ 
is generally superior to the Bayesian approach, especially when a 
specific metric for dealing with symbolic attributes is used.    
\end{abstract}

\section{INTRODUCTION}
\label{s-introduction}
Word Sense Disambiguation (\aWSD) is the problem of assigning 
the appropriate meaning (sense) to a given word in a text or 
discourse. 
Resolving the ambiguity of words is a central problem for language 
understanding applications and their associated tasks~\cite{ide98}, 
including, for instance, machine translation, information retrieval 
and hypertext navigation, parsing, speech synthesis, spelling
correction, reference resolution, automatic text summarization,
etc.
 
\aWSD\ is one of the most important open problems in the Natural Language
Processing (\aNLP) field. Despite the wide range of approaches
investigated and the large effort devoted to tackle this problem, 
it is a fact that to date no large--scale broad--coverage and highly 
accurate word sense disambiguation system has been built.

One of the most successful current lines of research is the corpus--based
approach in which statistical or Machine Learning (\aML) algorithms have 
been applied to learn statistical models or classifiers from corpora
in order to perform \aWSD. 
Generally, supervised approaches (those that learn from a previously 
semantically annotated corpus) have obtained better results than 
unsupervised methods on small sets of selected highly ambiguous
words, or artificial pseudo--words. Many standard \aML\ algorithms 
for supervised learning have been applied, such as: 
Bayesian learning~\cite{ng97a,pedersen98}, 
Exemplar--based learning \cite{ng96,ng97a,fujii98}, 
Decision Lists~\cite{yarowsky94},
Neural Networks~\cite{towell98}, etc.
Further, Mooney~\cite{mooney96} provides a comparative experiment 
on a very restricted domain between all previously cited methods but 
also including Decision Trees and Rule Induction algorithms.

Despite the good results obtained on limited domains, supervised 
methods suffer from the lack of widely available semantically 
tagged corpora, from which to construct really broad coverage systems.
This is known as the ``knowledge acquisition bottleneck''~\cite{gale93}.
Ng \cite{ng97b} estimates that the manual annotation effort necessary to 
build a broad coverage semantically annotated corpus would be about 16 
man-years. This extremely high overhead for supervision and,
additionally, the also serious learning overhead when common \aML\
algorithms scale to real size \aWSD\ problems,
explain why supervised methods have been seriously questioned.
 
Due to this fact, recent works have focused on 
reducing the acquisition cost as well as the need 
for supervision of corpus--based methods for \aWSD.
Consequently, the following three lines of research are currently
being studied: 
1) The design of
efficient example sampling methods~\cite{engelson96,fujii98}; 
2) The use of lexical resources, such as WordNet~\cite{miller90}, 
and WWW search engines to automatically obtain from Internet 
accurate and arbitrarily large word 
sense samples~\cite{leacock98,mihalcea99}; 
3) The use of unsupervised EM--like algorithms for estimating 
the statistical model parameters~\cite{pedersen98}.
It is our belief that this body of work, and in particular the
second line, provide enough evidence towards the ``opening'' of the
acquisition bottleneck in the near future.
For that reason, it is worth further investigating the 
application of supervised \aML\ methods to \aWSD, and 
thoroughly comparing existing alternatives. 

\subsection{Comments about Related Work}
Unfortunately, there have been very few direct comparisons between
alternative methods for \aWSD. However, it is commonly
stated that Naive Bayes, Neural Networks and Exemplar--based
learning represent state--of--the--art accuracy on supervised \aWSD
~\cite{mooney96,ng97a,leacock98,fujii98,pedersen98}. Regarding the 
comparison between Naive Bayes and Exemplar--based methods, 
the works by Mooney~\cite{mooney96} and Ng~\cite{ng97a} will be 
the ones basically referred to in this paper.

Mooney's paper shows that the Bayesian approach is clearly superior 
to the Exemplar--based approach.
Although it is not explicitly said, the overall accuracy of 
Naive Bayes is about 16 points higher than that of the 
Example--based algorithm, and the latter is only slightly above
the accuracy that a Most--Frequent--Sense classifier would obtain.  
In the Exemplar--based approach, the algorithm applied for classifying
new examples was a standard k-Nearest--Neighbour (\akNN), using the 
Hamming distance for measuring closeness. Neither example weighting 
nor attribute weighting are applied, $k$ is set to 3, and the number 
of attributes used is said to be almost 3,000.

The second paper compares the Naive Bayes approach with 
\aPEBLS~\cite{cost93}, a more sophisticated Exemplar--based 
learner especially designed for dealing with examples that 
have symbolic features. This paper shows that, for a large 
number of nearest--neighbours, the performance of both 
algorithms is comparable, while if cross validation 
is used for parameter setting, \aPEBLS\ slightly 
outperforms Naive Bayes.
It has to be noted that the comparison was carried out in a 
limited setting, using only 7 features, and that the 
attribute/example--weighting facilities provided by \aPEBLS\ 
were not used.
The author suggests that the poor results obtained 
in Mooney's work were due to the metric associated to 
the k-NN algorithm, but he did not test if the \aMVDM\ 
metric used in \aPEBLS\ is superior to the standard
Hamming distance or not.
 
Another surprising result that appears in Ng's paper is that the
accuracy results obtained were 1--1.6\% higher than those reported by the
same author one year before~\cite{ng96}, when running exactly the same
algorithm on the same data, but using a larger and richer set of attributes.
This apparently paradoxical difference is attributed, by the author, 
to the feature pruning process performed in the older paper. 

Apart from the contradictory results obtained by the previous
papers, some methodological drawbacks of both comparisons should also
be pointed out. On the one hand, Ng applies the algorithms 
on a broad--coverage corpus but reports the accuracy results of 
a single testing experiment, providing no statistical tests of 
significance.
On the other hand, Mooney performs thorough and rigorous experiments, 
but he compares the alternative methods on a limited domain 
consisting of a single word with a reduced set of six senses. 
Thus, it is our claim that this extremely specific domain does 
not guarantee the reaching of reliable conclusions about the 
relative performances of alternative methods when applied to 
broad--coverage domains.

Consequently, the aim of this paper is twofold: 1) To study the 
source of the differences between both approaches in order to clarify
the contradictory and incomplete information.
2) To empirically test the alternative algorithms and their extensions
on a broad--coverage sense tagged corpus, in order to estimate which 
is the most appropriate choice.

The paper is organized as follows: Section~\ref{s-methods} describes
the algorithms that will be tested, as well as the notation used. 
Section~\ref{s-setting} is devoted to carefully explain the
experimental setting. Section~\ref{s-experiments} reports the set of
experiments performed and the analysis of the results obtained.
The best alternative methods are tested on a broad coverage corpus in 
Section~\ref{s-global}. Finally, Section~\ref{s-conclusions} concludes 
and outlines some directions for future work.

\section{BASIC METHODS}
\label{s-methods}
\subsection{Naive Bayes}
The Naive Bayes classifier has been used in its most classical 
setting~\cite{duda73}. Let $C_{1} \ldots C_{m}$ the different classes 
and $\bigcap v_{j}$ the set of feature values of a test example. 
The Naive Bayes method tries to find the class that maximizes 
$P(C_{i}\mid \cap v_{j})$. 
Assuming independence between features, the goal of the algorithm can
be stated as:
$$\arg\max _{i}\: P(C_{i}\mid \cap v_{j})\,\approx\,\arg\max _{i}\: 
P(C_{i})\prod _{j}P(v_{j}\mid C_{i})\,,$$
\noindent where $P(C_{i}) \) and \( P(v_{j}\mid C_{i})$ are estimated 
during training process using relative frequencies.
To avoid the effects of zero counts when estimating the conditional 
probabilities of the model, a very simple smoothing technique,
proposed in Ng's paper~\cite{ng97a}, has been used. It consists in 
replacing zero counts of \( P(v_{j}\mid C_{i}) \) with $P(C_{i})/N$ 
where $N$ is the number of training examples.

Hereinafter, this method will be referred to as \aNB.

\subsection{Exemplar-Based Approach}
In our basic implementation all examples are stored in memory and the classification 
of a new example is based on a $k$--NN algorithm, which uses Hamming distance to 
measure closeness (in doing so, all examples are examined). If $k$ is greater than 1,
the resulting sense is the majority sense of the $k$ nearest neighbours.
Ties are resolved in favour of the most frequent sense among all those tied.
Hereinafter, this algorithm will be referred to as {\sf EB$_{h,k}$}.

In order to test some of the hypotheses about the differences between
Naive Bayes and Exemplar--based approaches, some variants of the basic
\akNN\ algorithm have been implemented:
\begin{itemize}
\item {\bf Example weighting}. This variant introduces a simple modification
  in the voting scheme of the $k$ nearest neighbours, which makes the 
  contribution of each example proportional to their importance.
  When classifying a new test example, each example of the set of 
  nearest neighbours votes for its class with a weight proportional 
  to its closeness to the test example. 
  Hereinafter, this variant will   be referred to as {\sf EB$_{h,k,e}$}.
\item {\bf Attribute weighting}. This variant consists of ranking all
  attributes by relevance and making them contribute to the distance 
  calculation with a weight proportional to their importance. The
  attribute weighting has been done using the \aRLM\ distance 
  measure~\cite{lopez91}. This measure, belonging to 
  the distance/information--based families of attribute selection
  functions, has been selected because it showed better performance 
  than seven other alternatives in an experiment of decision tree 
  induction for \aPOS\ tagging~\cite{marquez99}. 
  Hereinafter, this variant will be referred to as {\sf EB$_{h,k,a}$}.
\end{itemize}

When both modifications are put together, the resulting algorithm will be 
referred to as {\sf EB$_{h,k,e,a}$}. Finally, we have also
investigated the effect of using an alternative metric.

\begin{itemize}
\item {\bf Modified Value Difference Metric} (\aMVDM), proposed by Cost and
  Salzberg~\cite{cost93}, allows making graded guesses of the match
  between two different symbolic values. Let $v_{1}$ and $v_{2}$ be
  two values of a given attribute $a$. The \aMVDM\ distance between
  them is defined as:
  \[d(v_{1},v_{2})=\sum_{i=1}^m\, |P(C_i|v_1)-P(C_i|v_2)|\,
  \approx\sum_{i=1}^m\begin{array}{c}\left|\frac{N_{1,i}}{N_{1}}-\frac{N_{2,i}}
  {N_{2}}\right|\end{array}\]
  where $m$ is the number of classes, $N_{x,i}$ is the number of
  training examples with value $v_{x}$ of attribute $a$ that are
  classified as class $i$ in the training corpus and $N_{x}$ 
  is the number of training examples with value $v_{x}$ of attribute 
  $a$ in any class. Hereinafter, this variant will be referred to 
  as {\sf EB$_{cs,k}$}. This algorithm has also been used with the
  example--weighting facility ({\sf EB$_{cs,k,e}$}).
\end{itemize}

\section{SETTING}
\label{s-setting}
In our experiments, both approaches have been evaluated on the
\aDSO\ corpus, a semantically annotated corpus containing 
192,800 occurrences of 121 nouns and 70 verbs\footnote{These 
examples, consisting of the full sentence in which the ambiguous 
word appears, are tagged  with a set of labels corresponding, 
with minor changes, to the senses of WordNet 1.5~\cite{miller90}.}, 
corresponding to the most frequent and ambiguous English words. 
This corpus was collected by Ng and colleagues~\cite{ng96} and it is 
available from the Linguistic Data Consortium (LDC)\footnote{LDC 
address: {\tt http://www.ldc.upenn.edu/}}.

For our first experiments, a group of 15 words (10 nouns and 5 verbs) 
which frequently appear in the \aWSD\ literature has been 
selected. These words are described in the left hand--side of 
table~\ref{t-quinze}. 
Since our goal is to acquire a classifier for each word, each row 
represents a classification problem. The number of classes (senses) 
ranges from 4 to 30 and the number of training examples ranges from 373
to 1,500.
The \aMFS\ column of the table~\ref{t-quinze} show the percentage of
the most frequent sense for each word, i.e. the accuracy that a naive 
``Most--Frequent--Sense'' classifier would obtain. 

\begin{table}
\begin{center}
\caption{Set of 15 reference words.}
\vspace*{-3mm}
\label{t-quinze}
{\footnotesize\begin{tabular}{lccrr|cr} 
\hline
& & & & \multicolumn{1}{c}{\%} & \multicolumn{2}{c}{{\sf \# Attributes}} \\
\cline{5-7}
{\sf Word} & {\sf POS} & {\sf Sens.} &  {\sf Exs.} & {\sf MFS} & \aseta\ & \asetb\ \\
\hline
\ {\sf age}      & {\sf n} &  4 &  493 & 62.1 & 7 & 3,015\\
\ {\sf art}      & {\sf n} &  5 &  405 & 46.7 & 7 & 2,641\\
\ {\sf car}      & {\sf n} &  5 & 1,381 & 95.1 & 7 & 4,719\\
\ {\sf child}    & {\sf n} &  4 & 1,068 & 80.9 & 7 & 4,840\\
\ {\sf church}   & {\sf n} &  4 &  373 & 63.1 & 7 & 2,375\\
\ {\sf cost}     & {\sf n} &  3 & 1,500 & 87.3 & 7 & 4,930\\
\ {\sf fall}     & {\sf v} & 19 & 1,500 & 70.1 & 7 & 4,173\\
\ {\sf head}     & {\sf n} & 14 &  870 & 36.9 & 7 & 4,284\\
\ {\sf interest} & {\sf n} &  7 & 1,500 & 45.1 & 7 & 5,328\\
\ {\sf know}     & {\sf v} &  8 & 1,500 & 34.9 & 7 & 5,301\\
\ {\sf line}     & {\sf n} & 26 & 1,342 & 21.9 & 7 & 5,813\\
\ {\sf set}      & {\sf v} & 19 & 1,311 & 36.9 & 7 & 5,749\\
\ {\sf speak}    & {\sf v} &  5 &  517 & 69.1 & 7 & 2,975\\
\ {\sf take}     & {\sf v} & 30 & 1,500 & 35.6 & 7 & 6,428\\
\ {\sf work}     & {\sf n} &  7 & 1,469 & 31.7 & 7 & 6,321\\
\hline
\multicolumn{2}{l}{\ Avg. nouns}           &  8.6 & 1,040.1 & 57.4 & 7 & 4,935.0\\
\multicolumn{2}{l}{\ \phantom{Avg.} verbs} & 17.9 & 1,265.6 & 46.6 & 7 & 5,203.5\\
\hline
\multicolumn{2}{l}{\ \phantom{Avg.} all}   & 12.1 & 1,115.3 & 53.3 & 7 & 5,036.6\\
\hline
\end{tabular}}
\vspace*{-2mm}
\end{center}
\end{table}
%
\noindent Two sets of attributes have been used, which 
will be referred to as \aseta\ and \asetb, respectively.
Let ``$\dots\, w_{-3}$ $w_{-2}$ $w_{-1}$ $w$ $w_{+1}$ $w_{+2}$
$w_{+3}\dots$'' be the context of consecutive words around the word 
$w$ to be disambiguated. Attributes refer to this context as follows.
\begin{itemize}
\item \aseta\ contains the seven following attributes: $w_{-2}$, 
$w_{-1}$, $w_{+1}$, $w_{+2}$, $(w_{-2},w_{-1})$, ($w_{-1},w_{+1}$), 
and $(w_{+1},w_{+2})$, where the last three correspond to collocations 
of two consecutive words. 
These attributes, which are exactly those used 
in~\cite{ng97a}, represent the {\it local context} of the ambiguous
word and they have been proven to be very informative features for \aWSD.
Note that whenever an attribute refers to a position
that falls beyond the boundaries of the sentence for a certain
example, a default value ``\_$\!$\_'' is assigned.
\end{itemize}
Let $p_{\pm i}$ be the part--of--speech tag of word $w_{\pm i}$, and  
$c_{1}, \ldots, c_{m}$ the unordered set of open class words
appearing in the sentence. 
\begin{itemize}
\item \asetb\ enriches the local context: $w_{-1}$, $w_{+1}$, 
$(w_{-2},w_{-1})$, ($w_{-1},w_{+1}$), $(w_{+1},w_{+2})$, 
$(w_{-3}, w_{-2}, w_{-1})$, $(w_{-2}, w_{-1}, w_{+1})$, 
$(w_{-1}, w_{+1}, w_{+2})$ and $(w_{+1}, w_{+2}, w_{+3})$, 
with the part--of--speech information: $p_{-3}$, $p_{-2}$, 
$p_{-1}$, $p_{+1}$, $p_{+2}$, $p_{+3}$, and, additionally, 
it incorporates {\it broad context} information: $c_{1} \ldots c_{m}$.
\asetb\ is intended to represent a more realistic set of attributes
for \aWSD\footnote{In fact, it incorporates all the attributes used 
in~\cite{ng96}, with the exception of the morphology of the target 
word and the verb--object syntactic relation.}.
Note that $c_i$ attributes are binary--valued, denoting the
the presence or absence of a content word in the sentence context.
\end{itemize}

The right hand--side of table~\ref{t-quinze} contains the information 
about the number of features. Note that \aseta\ has a constant 
number of attributes (7), while for \asetb\ this number depends on the 
concrete word, and that it ranges from 2,641 to 6,428. 

\section{EXPERIMENTS}
\label{s-experiments}
The comparison of algorithms has been performed in series 
of controlled experiments using exactly the same training and 
test sets for each method. 
The experimental methodology consisted on a 10-fold cross-validation.
All accuracy/error rate figures appearing in the paper are averaged 
over the results  of the 10 folds. The statistical tests of 
significance have been performed using a 10-fold cross validation 
paired Student's $t$-test~\cite{dietterich98} with a confidence value of: 
$t_{9,0.975}=2.262$.

Exemplar-based algorithms are run several times using different number
of nearest neighbours (1, 3, 5, 7, 10, 15, 20 and 25) and the results
corresponding to the best choice are reported\footnote{In order to
  construct a real \akNN--based system for \aWSD, the 
  $k$ parameter should be estimated by cross--validation using only 
  the training set~\cite{ng97a}, however, in our case, this 
  cross--validation inside the cross--validation involved in the testing
  process would generate a prohibitive overhead.}.

\subsection{Using \aseta}
Table~\ref{t-ldcng97} shows the results of all methods and variants 
tested on the 15 reference words, using the \aseta\ set of
attributes: Most Frequent Sense (\aMFS), 
Naive Bayes (\aNB), Exemplar--based using Hamming
distance (\aEBh\ variants, 5th to 9th columns), and Exemplar-based
approach using the \aMVDM\ metric (\aEBcs\ variants, 10th to 12th 
columns) are included.
The best result for each word is printed in boldface.
\begin{table*}
\begin{center}
\caption{Results of all algorithms on the set of 15 reference words using \aseta.}
\vspace*{-3mm}
\label{t-ldcng97}
{\footnotesize\begin{tabular}{lcc|c|ccccc|ccc}
\hline
 & & \multicolumn{10}{c}{{\sf Accuracy (\%)}} \\
\cline{3-12}
{\sf Word} & {\sf POS} & {\sf MFS} & {\sf NB} & {\sf EB$_{h,1}$} & {\sf EB$_{h,7}$} & {\sf EB$_{h,15,e}$} & {\sf EB$_{h,7,a}$} & {\sf EB$_{h,7,e,a}$} & {\sf EB$_{cs,1}$} & {\sf EB$_{cs,10}$} & {\sf EB$_{cs,10,e}$} \\
\hline
\ {\sf age}      & {\sf n} & 62.1 & 73.8 & 71.4 & 69.4 & 71.0 & 74.4 & {\bf 75.9} & 70.8 & 73.6 & 73.6 \\
\ {\sf art}      & {\sf n} & 46.7 & 54.8 & 44.2 & 59.3 & 58.3 & 58.5 & 57.0 & 54.1 & 59.5 & {\bf 61.0} \\
\ {\sf car}      & {\sf n} & 95.1 & 95.4 & 91.3 & 95.5 & 95.8 & 96.3 & 96.2 & 95.4 & {\bf 96.8} & {\bf 96.8} \\
\ {\sf child}    & {\sf n} & 80.9 & 86.8 & 82.3 & 89.3 & 89.5 & 91.0 & {\bf 91.2} & 87.5 & 91.0 & 90.9 \\
\ {\sf church}   & {\sf n} & 61.1 & 62.7 & 61.9 & 62.7 & 63.0 & 62.5 & 64.1 & 61.7 & {\bf 64.6} & 64.3 \\
\ {\sf cost}     & {\sf n} & 87.3 & 86.7 & 81.1 & 87.9 & 87.7 & {\bf 88.1} & 87.8 & 82.5 & 85.4 & 84.7 \\
\ {\sf fall}     & {\sf v} & 70.1 & 76.5 & 73.3 & 78.2 & 79.0 & 78.1 & 79.8 & 78.7 & 81.6 & {\bf 81.9} \\
\ {\sf head}     & {\sf n} & 36.9 & 76.9 & 70.0 & 76.5 & 76.9 & 77.0 & 78.7 & 74.3 & 78.6 & {\bf 79.1} \\
\ {\sf interest} & {\sf n} & 45.1 & 64.5 & 58.3 & 62.4 & 63.3 & 64.8 & 66.1 & 65.1 & 67.3 & {\bf 67.4} \\
\ {\sf know}     & {\sf v} & 34.9 & 47.3 & 42.2 & 44.3 & 46.7 & 44.9 & 46.8 & 45.1 & 49.7 & {\bf 50.1} \\
\ {\sf line}     & {\sf n} & 21.9 & 51.9 & 46.1 & 47.1 & 49.7 & 50.7 & 51.9 & 53.3 & {\bf 57.0} & 56.9 \\
\ {\sf set}      & {\sf v} & 36.9 & 55.8 & 43.9 & 53.0 & 54.8 & 52.3 & 54.3 & 49.7 & {\bf 56.2} & 56.0 \\
\ {\sf speak}    & {\sf v} & 69.1 & {\bf 74.3} & 64.6 & 72.2 & 73.7 & 71.8 & 72.9 & 67.1 & 72.5 & 72.9 \\
\ {\sf take}     & {\sf v} & 35.6 & 44.8 & 39.3 & 43.7 & 46.1 & 44.5 & 46.0 & 45.3 & 48.8 & {\bf 49.1} \\
\ {\sf work}     & {\sf n} & 31.7 & 51.9 & 42.5 & 43.7 & 47.2 & 48.5 & 48.9 & 48.5 & 52.0 & {\bf 52.5} \\
\hline
\multicolumn{2}{l}{\ Avg. nouns}           & 57.4 & 71.7 & 65.8 & 70.0 & 71.1 & 72.1 & 72.6 & 70.6 & 73.6 & {\bf 73.7} \\
\multicolumn{2}{l}{\ \phantom{Avg.} verbs} & 46.6 & 57.6 & 51.1 & 56.3 & 58.1 & 56.4 & 58.1 & 55.9 & 60.3 & {\bf 60.5} \\
\hline
\multicolumn{2}{l}{\ \phantom{Avg.} all}   & 53.3 & 66.4 & 60.2 & 64.8 & 66.2 & 66.1 & 67.2 & 65.0 & 68.6 & {\bf 68.7} \\
\hline
\end{tabular}}
\end{center}
\end{table*}
From these figures, several conclusions can be drawn:
\begin{itemize}
\item All methods significantly outperform the \aMFS\ classifier.
\item Referring to the \aEBh\ variants, \aEBhvii\ performs
  significantly better than \aEBhi, confirming the results of
  Ng~\cite{ng97a} that values of $k$ greater than one are needed in order 
  to achieve good performance with the \akNN\ approach.
  Additionally, both example weighting (\aEBhxve) and attribute weighting  
  (\aEBhviia) significantly improve \aEBhvii. Further, the combination 
  of both (\aEBhviiea) achieves an additional improvement.
\item The \aMVDM\ metric is superior to Hamming distance. 
  The accuracy of \aEBcsxe\ is significantly higher than those
  of any \aEBh\ variant. Unfortunately, the use of weighted examples
  does not lead to further improvement in this case. A drawback of
  using the \aMVDM\ metric is the computational overhead introduced by
  its calculation. Table~\ref{t-time15} shows that \aEBh\ is fifty
  times faster than \aEBcs\ using \aseta\footnote{The current
  programs are implemented using PERL-5.003 and they run on 
  a {\sc Sun} UltraSPARC-2 machine with 192Mb of RAM.}.
\item The Exemplar-based approach achieves better 
    results than the Naive Bayes algorithm. This difference is
    statistically significant when comparing the \aEBcsx\ and
    \aEBcsxe\ against \aNB.
\end{itemize}
%
\subsection{Using \asetb}
The aim of the experiments with \asetb\ is to test both methods 
with a realistic large set of features. 
Table~\ref{t-ldcng96} summarizes the results of these
experiments\footnote{Detailed results for each word are not included.}.
\begin{table*}
\begin{center}
\caption{Results of all algorithms on the set of 15 reference words using \asetb.}
\vspace*{-3mm}
\label{t-ldcng96}
{\footnotesize\begin{tabular}{lc|cc|cccccc|ccc} 
\hline
& \multicolumn{12}{c}{{\sf Accuracy (\%)}} \\
\cline{2-13}
{\sf POS} & {\sf MFS} & {\sf NB} & {\sf PNB} & {\sf EB$_{h,15}$} & {\sf PEB$_{h,1}$} & {\sf PEB$_{h,7}$} & {\sf PEB$_{h,7,e}$} & {\sf PEB$_{h,7,a}$} & {\sf PEB$_{h,10,e,a}$} & {\sf PEB$_{cs,1}$} & {\sf PEB$_{cs,10}$} & {\sf PEB$_{cs,10,e}$} \\
\hline
nouns  & 57.4 & 72.2 & 72.4 & 64.3 & 70.6 & 72.4 & 73.7 & 72.5 &
73.4 & 73.2 & 75.4 & {\bf 75.6} \\
verbs  & 46.6 & 55.2 & 55.3 & 43.0 & 54.7 & 57.7 & 59.5 & 58.9 & 60.2
& 58.6 & 61.9 & {\bf 62.1}\\
\hline
all    & 53.3 & 65.8 & 66.0 & 56.2 & 64.6 & 66.8 & {\bf 68.4} & 67.4 &
68.4 & 67.7 & 70.3 & {\bf 70.5}\\
\hline
\end{tabular}}
\end{center}
\end{table*}

Let's now consider only \aNB\ and \aEBh\ (3rd and 5th columns).
A very surprising result is observed: while \aNB\ achieves 
almost the same accuracy that in the previous experiment, the 
exemplar--based approach shows a very low performance. The 
accuracy of \aEBh\ drops 8.6 points (from 6th column of 
table~\ref{t-ldcng97} to 5th column of table~\ref{t-ldcng96}) 
and is only slightly higher than that of \aMFS. 

The problem is that the binary representation of the broad--context
attributes is not appropriate for the \akNN\ algorithm. Such a 
representation leads to an extremely sparse vector representation 
of the examples, since in each example only a few words, among 
all possible, are observed. Thus, the examples are represented by a 
vector of about 5,000 0's and only a few 1's. 
In this situation two examples will coincide in the majority of the values
of the attributes (roughly speaking in ``all'' the zeros) and will 
probably differ in those positions corresponding to 1's.
This fact wrongly biases the similarity measure (and thus the
classification) in favour of that stored examples which have less 
1's, that is, those corresponding to the shortest sentences.

This situation could explain the poor results obtained by the \akNN\
algorithm in Mooney's work, in which a large number of attributes was
used. Further, it could explain why the results of Ng's system working
with a rich attribute set (including binary--valued contextual
features) were lower than those obtained with a simpler set of
attributes\footnote{Recall that authors attributed the bad results to
  the absence of attribute weighting and to the attribute pruning, 
respectively.}.

In order to address this limitation we propose to reduce the attribute
space by collapsing all binary attributes $c_{1},\ldots, c_{m}$ in a 
single set--valued attribute $c$ that contains, for each example,  all
content words that appear in the sentence. In this setting, the 
similarity $S$ between two values $V_i=\{w_{i_1},w_{i_2},\dots,w_{i_n}\}$ 
and $V_j=\{w_{j_1},w_{j_2},\dots,w_{j_m}\}$ can be redefined as: 
$S(V_i,V_j)=|\!|\,V_i\cap V_j\,|\!|$, that is, equal to the number of words
shared\footnote{This measure is usually known as the {\it matching
coefficient}~\cite{manning99}. More complex similarity measures, 
e.g. Jaccard or Dice coefficients, have not been explored.}.

This approach implies that a test example is classified taking 
into account the information about the words it contains 
({\it positive} information), but no the information about the 
words it does not contain. Besides, it allows a very efficient 
implementation, which will be referred to as \aPEB\ (standing for
Positive Exemplar--Based).

In the same direction, we have tested the Naive Bayes algorithm
combining only the conditional probabilities corresponding to the
words that appear in the test examples. This variant is referred to as
\aPNB. The results of both \aPEB\ and \aPNB\ are included in 
table~\ref{t-ldcng96}, from which the following conclusions can 
be drawn.
 
\begin{itemize}
\item The \aPEB\ approach reaches excellent results, improving by 
  10.6 points the accuracy of \aEB\ (see 5th and 7th columns of
  table~\ref{t-ldcng96}). Further, the results obtained
  significantly outperform those obtained using \aseta,   
  indicating that the (careful) addition of richer attributes 
  leads to more accurate classifiers.   
  Additionally, the behaviour of the different variants is similar to
  that observed when using \aseta, with the exception that the
  addition of attribute--weighting to the example--weighting (\aPEBhxea)
  seems no longer useful.
\item \aPNB\ algorithm is at least as accurate as \aNB.
\item Table~\ref{t-time15} shows that the {\it positive} approach 
  increases greatly the efficiency of the algorithms. The acceleration
  factor is 80 for \aNB\ and 15 for \aEBh\ (the calculation of \aEBcs\
  variants was simply not feasible working with the attributes of \asetb).
\item The comparative conclusions between the Bayesian and
  Exemplar--based approaches reached in the experiments using
  \aseta\ also hold here. Further, the accuracy of \aPEBhviie\ is now
  significantly higher than that of \aPNB. 
\end{itemize}
\vspace*{-4mm}
\begin{table}
\begin{center}
\caption{CPU--time elapsed on the set of 15 words (``hh:mm'').}
\vspace*{-3mm}
\label{t-time15}
{\footnotesize\begin{tabular}{lcccccc}
\hline\rule{0pt}{12pt}
& \aNB & & \aEBhxve & & \aEBhviia & \aEBcsxe\\\hline
\aseta & 00:07 & & 00:08 & & 00:11 & 09:56\\
\hline\rule{0pt}{12pt}
& \aNB & \aPNB & \aEBhxve & \aPEBhviie & \aPEBhviia & \aPEBcsxe\\\hline
\asetb & 16:13 & 00:12 & 06:04 & 00:25 & 03:55 & 49:43\\
\hline
\end{tabular}}
\end{center}
\vspace*{-6mm}
\end{table}

\section{GLOBAL RESULTS}
\label{s-global}
In order to ensure that the results obtained so far also hold on 
a realistic broad--coverage domain, the \aPNB\ and \aPEB\ algorithms have been 
tested on the whole sense--tagged corpus, using both sets of attributes. 
This corpus contains about 192,800 examples of 121 nouns and 
70 verbs. The average number of senses is 7.2 for nouns, 12.6 for verbs, 
and 9.2 overall. The average number of training examples is 933.9 
for nouns, 938.7 for verbs, and 935.6 overall. 

The results obtained are presented in table~\ref{t-ldc191}.
It has to be noted that the results of \aPEBcs\ using \asetb\ 
were not calculated due to the extremely large computational 
effort required by the algorithm (see table~\ref{t-time15}).
Results are coherent to those reported previously, that is:
\setlength{\tabcolsep}{1.3mm}
\begin{table}
\begin{center}
\caption{Global results on the 191--word corpus.}
\vspace*{-3mm}
\label{t-ldc191}
{\footnotesize\begin{tabular}{l|l|cccc|ccc}
\hline
& & \multicolumn{4}{c}{{\sf Accuracy (\%)}} & \multicolumn{3}{c}{{\sf CPU--Time (hh:mm)}} \\
\cline{3-6}
\cline{7-9}
& {\sf POS} & {\sf MFS} & {\sf PNB} & {\sf PEB$_{h}$} & {\sf PEB$_{cs}$} & {\sf PNB} & {\sf PEB$_{h}$} & {\sf PEB$_{cs}$} \\
\hline
       & nouns  & 56.4 & 68.7 & 68.5 & {\bf 70.2}\\
\aseta & verbs  & 48.7 & 64.8 & 65.3 & {\bf 66.4} & 00:33 & 00:47 & 92:22 \\
       & all    & 53.2 & 67.1 & 67.2 & {\bf 68.6}\\
\hline
       & nouns  & 56.4 & 69.2 & {\bf 70.1} &\\
\asetb & verbs  & 48.7 & 63.4 & {\bf 67.0} & -- & 01:06 & 01:46 & -- \\
       & all    & 53.2 & 66.8 & {\bf 68.8} &\\
\hline
\end{tabular}}
\end{center}
\end{table}
\vspace*{-6mm}
\begin{itemize}
\item In \aseta, the Exemplar--based approach using the \aMVDM\ metric is
  significantly superior to the rest.
\item In \asetb, the Exemplar--based approach using Hamming distance
  and example weighting significantly outperforms the Bayesian
  approach. Although the use of the \aMVDM\ metric could lead to
  better results, the current implementation is computationally 
  prohibitive.   
\item Contrary to the Exemplar-based approach, Naive Bayes
  does not improve accuracy when moving from \aseta\ to \asetb,
  that is, the simple addition of attributes does not guarantee
  accuracy improvements in the Bayesian framework.   
\end{itemize}

\section{CONCLUSIONS}
\label{s-conclusions}
This work has focused on clarifying some contradictory results obtained
when comparing Naive Bayes and Exemplar--based approaches to \aWSD. 
Different alternative algorithms have been tested using two 
different attribute sets on a large sense--tagged corpus. 
The experiments carried out show that
Exemplar--based algorithms have generally better performance than 
Naive Bayes, when they are extended with example/attribute weighting,
richer metrics, etc.

The reported experiments also show that the Exemplar--based approach 
is very sensitive to the representation of a concrete type of 
attributes, frequently used in Natural Language problems.
To avoid this drawback, an alternative representation of the
attributes has been proposed and successfully tested. 
Furthermore, this representation also improves the efficiency 
of the algorithms, when using a large set of attributes.

The test on the whole corpus allows us to estimate that, in a
realistic scenario, the best tradeoff between performance and
computational requirements is achieved by using the Positive 
Exemplar--based algorithm,  \asetb\ set of attributes, Hamming
distance, and example--weighting.

Further research on the presented algorithms to be 
carried out in the near future includes: 1) The study of the 
behaviour with respect to the number of training examples; 
2) The study of the robustness in the presence of highly 
redundant attributes; 
3) The testing of the algorithms on alternative sense--tagged 
corpora automatically acquired from Internet.

\ack This research has been partially funded by the Spanish Research
Department (CICYT's project TIC98--0423--C06) and by the
Catalan Research Department (CIRIT's consolidated research group
1999SGR-150, CREL's Catalan WordNet project and CIRIT's grant 1999FI
00773).

We would also like to thank the referees for their valuable comments.
%
\bibliography{../bibliografia/bibliografia}
\end{document}